\documentclass{article}

\usepackage{microtype}
\usepackage{graphicx}
\usepackage{subcaption}
\usepackage{booktabs} 

\usepackage{hyperref}
\usepackage{svg}

\usepackage[preprint]{icml2026}

\usepackage{amsmath}
\usepackage{amssymb}
\usepackage{mathtools}
\usepackage{amsthm}

\usepackage[capitalize,noabbrev]{cleveref}

\theoremstyle{plain}

\theoremstyle{definition}

\theoremstyle{remark}

\usepackage[textsize=tiny]{todonotes}

\icmltitlerunning{Ref-SAM3D: Bridging SAM3D with Text for Reference 3D Reconstruction}

\begin{document}

\twocolumn[
  \icmltitle{Ref-SAM3D: Bridging SAM3D with Text for Reference 3D Reconstruction}



  \icmlsetsymbol{equal}{*}

  \begin{icmlauthorlist}
    \icmlauthor{Yun Zhou}{equal,yyy}
    \icmlauthor{Yaoting Wang}{equal,yyy}
    \icmlauthor{Guangquan Jie}{yyy}
    \icmlauthor{Jinyu Liu}{yyy}
    \icmlauthor{Henghui Ding}{yyy}
  \end{icmlauthorlist}

  \icmlaffiliation{yyy}{Institute of Big Data, College of Computer Science and Artificial Intelligence, Fudan University}

  \icmlcorrespondingauthor{Henghui Ding}{hhding@fudan.edu.cn}

  \icmlkeywords{Machine Learning, ICML}

  \vskip 0.3in
]



\printAffiliationsAndNotice{}  

\begin{abstract}
SAM3D has garnered widespread attention for its strong 3D object reconstruction capabilities. However, a key limitation remains: SAM3D cannot reconstruct specific objects referred to by textual descriptions, a capability that is essential for practical applications such as 3D editing, game development, and virtual environments. To address this gap, we introduce \textbf{Ref-SAM3D}, a simple yet effective extension to SAM3D that incorporates textual descriptions as a high-level prior, enabling text-guided 3D reconstruction from a single RGB image. Through extensive qualitative experiments, we show that Ref-SAM3D, guided only by natural language and a single 2D view, delivers competitive and high-fidelity zero-shot reconstruction performance. Our results demonstrate that Ref-SAM3D effectively bridges the gap between 2D visual cues and 3D geometric understanding, offering a more flexible and accessible paradigm for reference-guided 3D reconstruction. Code is available at: \url{https://github.com/FudanCVL/Ref-SAM3D}.
\end{abstract}

\section{Introduction}
Recent advances in foundation models have revolutionized a wide range of 2D vision tasks, with the Segment Anything Model (SAM) \cite{sam,sam2,sam3} emerging as a particularly powerful framework for open-world segmentation driven by spatial prompts such as points, bounding boxes, or masks. By providing strong and generalizable segmentation priors, SAM series have reshaped how users interact with images, enabling flexible object extraction without task-specific training. This paradigm has further inspired progress in the 3D domain, most notably through SAM3D, which extends SAM’s segmentation capabilities to 3D understanding. SAM3D uses segmentation masks as spatial cues to guide single-image 3D reconstruction, allowing users to specify objects of interest and obtain corresponding 3D representations with minimal effort.

Despite its strong performance, current SAM3D remains fundamentally limited by its reliance on spatial references (e.g., user-provided masks). In many real-world scenarios, supplying precise masks is inconvenient, ambiguous, or even infeasible, particularly when the target object is best identified through high-level semantic attributes rather than explicit spatial localization. Such limitations motivate the need for a more natural and accessible form of interaction, where users can specify reconstruction targets simply by describing them in language.

In this work, we introduce \textbf{Ref-SAM3D}, a simple yet effective extension that augments SAM3D with natural language as an alternative high-level reference signal. By integrating a mask proposer equipped with vision-language grounding capabilities, Ref-SAM3D enables users to reconstruct 3D objects solely through natural-language descriptions, without requiring any explicit spatial input. This design allows the system to interpret semantic cues from text and automatically produce the corresponding object masks, which are then used by SAM3D for object-centric reconstruction. In doing so, our approach unifies textual and spatial references within a common reference-guided 3D object reconstruction framework, significantly enhancing accessibility and interaction flexibility while preserving SAM3D’s strong reconstruction quality.

The proposed Ref-SAM3D introduces minimal architectural changes and requires no retraining of either the base mask proposer or the SAM3D object reconstructor, making Ref-SAM3D highly practical, modular, and plug-and-play with existing workflows. Through extensive qualitative experiments, we demonstrate that Ref-SAM3D achieves competitive and robust zero-shot 3D reconstruction performance using only natural-language references. These results validate the effectiveness of our extension and highlight the potential of language-guided 3D object reconstruction from a single image.

We hope that Ref-SAM3D will serve as a stepping stone toward more intuitive, language-driven 3D content creation from common and massive 2D images, inspiring future research on multimodal 3D understanding and enabling broader accessibility in 3D editing, content creation, and interactive digital environments.

\section{Related Work}

\subsection{3D Reconstruction}
Recent advancements in 3D reconstruction have demonstrated the wide range of applications and the powerful capabilities of modern methods in generating high-quality 3D objects. Leveraging the expressive power of diffusion models, SDFusion \cite{sdfusion} enables multimodal 3D shape generation from images, text, and partial shapes, unifying tasks such as shape completion and text-to-3D within a single framework. Building on the success of vision–language models and foundation models like SAM, Anything-3D \cite{anything3d} achieves zero-shot, category-agnostic 3D reconstruction from a single RGB image without requiring task-specific training. Further integrating large language models, ShapeGPT \cite{shapegpt} formulates 3D shapes as a language-like sequence, supporting instruction-driven generation and enabling versatile multimodal interactions across shape editing, completion, and cross-modal translation. More recently, efforts have shifted toward structural and geometric faithfulness: Part123 \cite{part123} introduces part-aware 3D reconstruction by combining multiview-consistent diffusion with SAM-derived segmentation masks and contrastive learning, producing meshes with semantically meaningful part decompositions that facilitate downstream tasks like primitive fitting and shape editing. Complementing this direction, LAM3D \cite{lam3d} proposes an Image–Point–Cloud Feature Alignment mechanism that efficiently maps single-image features to latent tri-planes derived from 3D point clouds, yielding high-fidelity meshes with reduced geometric distortion in under 6 seconds. Meanwhile, MTFusion \cite{mtfusion} enhances surface detail and training efficiency by coupling text–image multimodal priors with a specialized SDF decoder based on FlexiCubes, enabling high-fidelity reconstruction guided by both visual and textual cues. Collectively, these works illustrate the rapid progress toward general-purpose, semantically rich, and geometry-accurate 3D reconstruction from minimal inputs.

\subsection{Segment Anything Model}
The Segment Anything Model (SAM) series proposed aims to build a general-purpose and promptable foundation model for segmentation, and has rapidly reshaped the landscape of visual perception. SAM~1 \cite{sam}, trained on the large-scale SA-1B dataset, supports various spatial prompts including points, boxes, and masks, enabling strong zero-shot interactive segmentation without task-specific fine-tuning. SAM~2 \cite{sam2} extends this paradigm from static images to videos, enabling long-term object tracking with efficient interaction. Most recently, SAM~3 \cite{sam3} expands the prompt space to high-level concepts and incorporates textual phrases, achieving multi-modal concept segmentation across images and videos and demonstrating significant improvements over SAM~2 on the challenging MOSEv2 dataset~\cite{MOSEv2}.

Owing to the generality and strong segmentation priors, the SAM family has become the default backbone for numerous downstream tasks, inspiring a broad range of extensions. In text-vision scenarios, Grounded-SAM combines open-vocabulary detection with SAM for text-driven open-world segmentation. In medical imaging, MedSAM \cite{medsam} transfers SAM’s capabilities to CT/MRI data. In audio-visual understanding, GAVS \cite{gavs} encodes audio signals as semantic prompts to guide SAM toward sounding objects. In remote sensing, SAM-RS \cite{samrs} explores the adaptability of SAM in remote sensing with multi-sclae datasets. These works collectively demonstrate that SAM serves as a powerful and versatile base model that can be effectively adapted through prompting and lightweight modification~\cite{MeViSv2,ding2025multimodal}.

With 2D segmentation capabilities becoming increasingly mature, SAM 3D \cite{sam3d} extends the SAM family to single-image 3D reconstruction. By leveraging high-quality 2D segmentation priors together with large-scale image-3D data, SAM 3D achieves state-of-the-art performance for both general objects and human bodies. However, current SAM 3D pipelines require foreground masks provided via user interaction, which limits their usability in scenarios requiring automation or accessibility. Therefore, shifting from a reliance on explicit mask prompts to a more flexible interaction in 3D reconstruction is a promising research direction worth exploring.

\section{Method}
\label{sec:method}

\begin{figure*}[t]
  \begin{center}
    \includegraphics[width=1\textwidth]{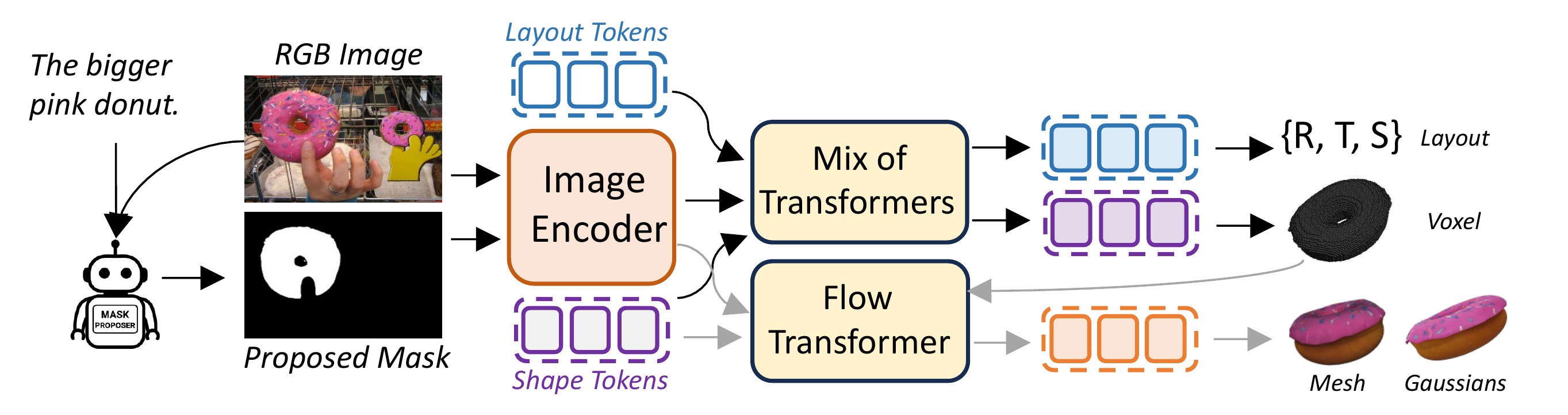}
    \vspace{-3mm}
    \caption{
        Inference pipeline of Ref-SAM3D. The pipeline takes an input image and a referring expression, which are processed by the mask proposer to generate the mask of the referred object. This mask is then passed to SAM3D with the original image for 3D object reconstruction. For simplicity, the layout, voxel, mesh, and Gaussian splat decoders are omitted. The output ${R, T, S}$ represents the layout attributes, including rotation, translation, and scaling.
    }
    \label{fig:pipeline}
  \end{center}
\end{figure*}

\begin{figure*}[p]
  \begin{center}
    \includegraphics[width=0.96\textwidth]{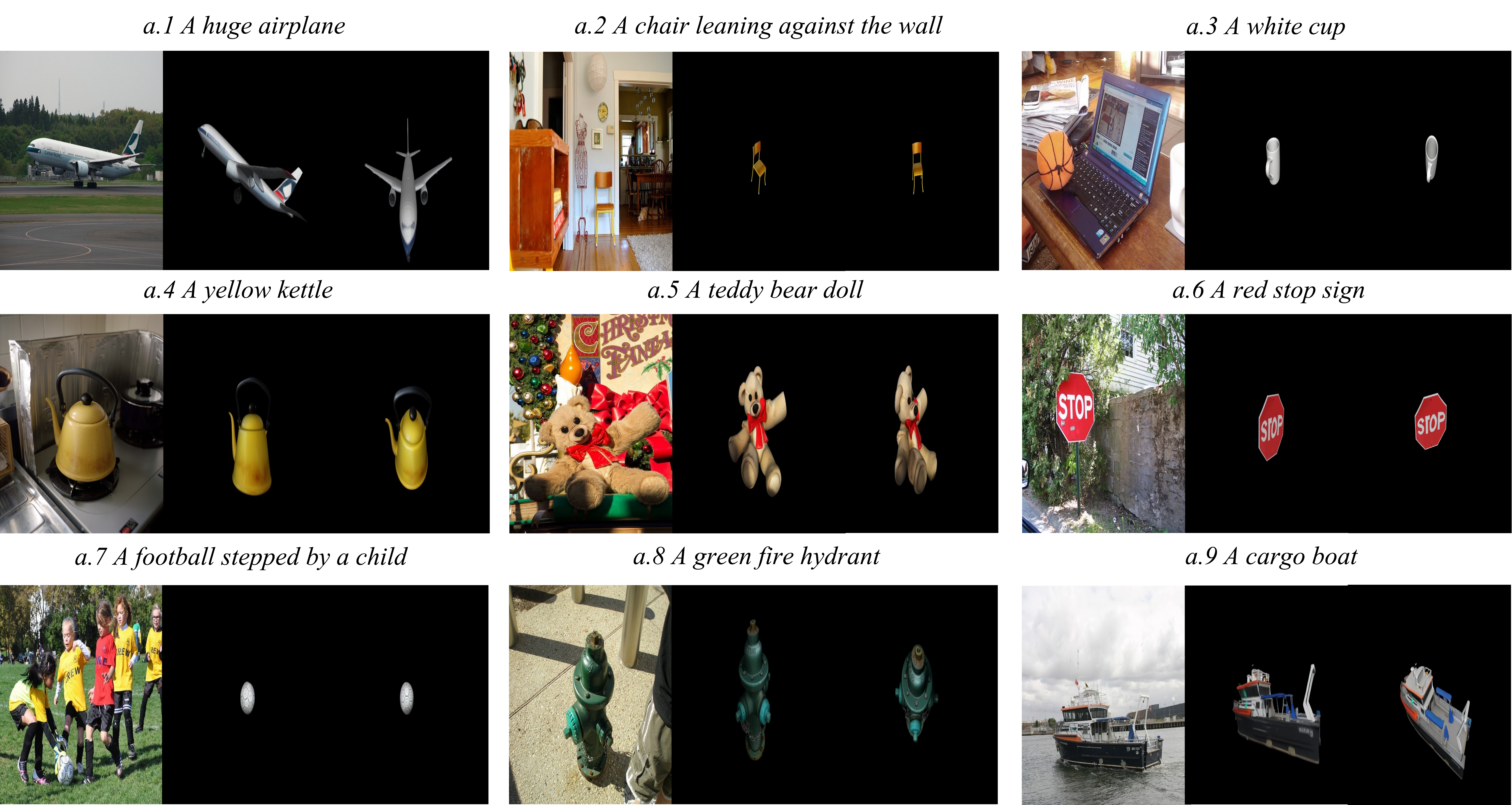}
    \caption{Case A: Referring to and reconstructing a single object in straightforward scenarios.}
    \label{fig:case_a}
  \end{center}
\end{figure*}
\begin{figure*}[p]
  \begin{center}
    \includegraphics[width=\textwidth]{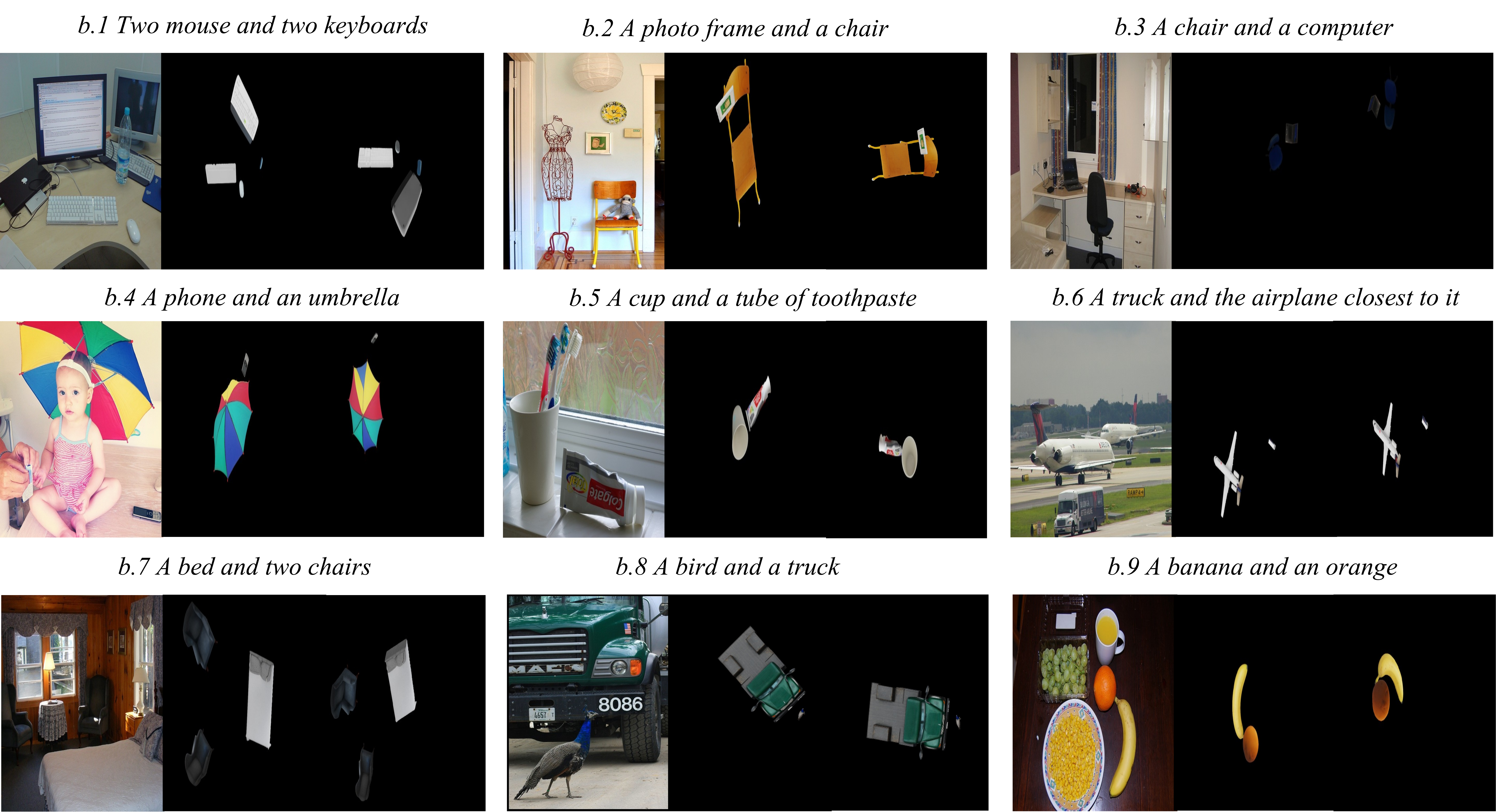}
    \caption{Case B: Referring to and reconstructing multiple objects.}
    \label{fig:case_b}
  \end{center}
\end{figure*}

\begin{figure*}[t]
  \begin{center}
    \includegraphics[width=\textwidth]{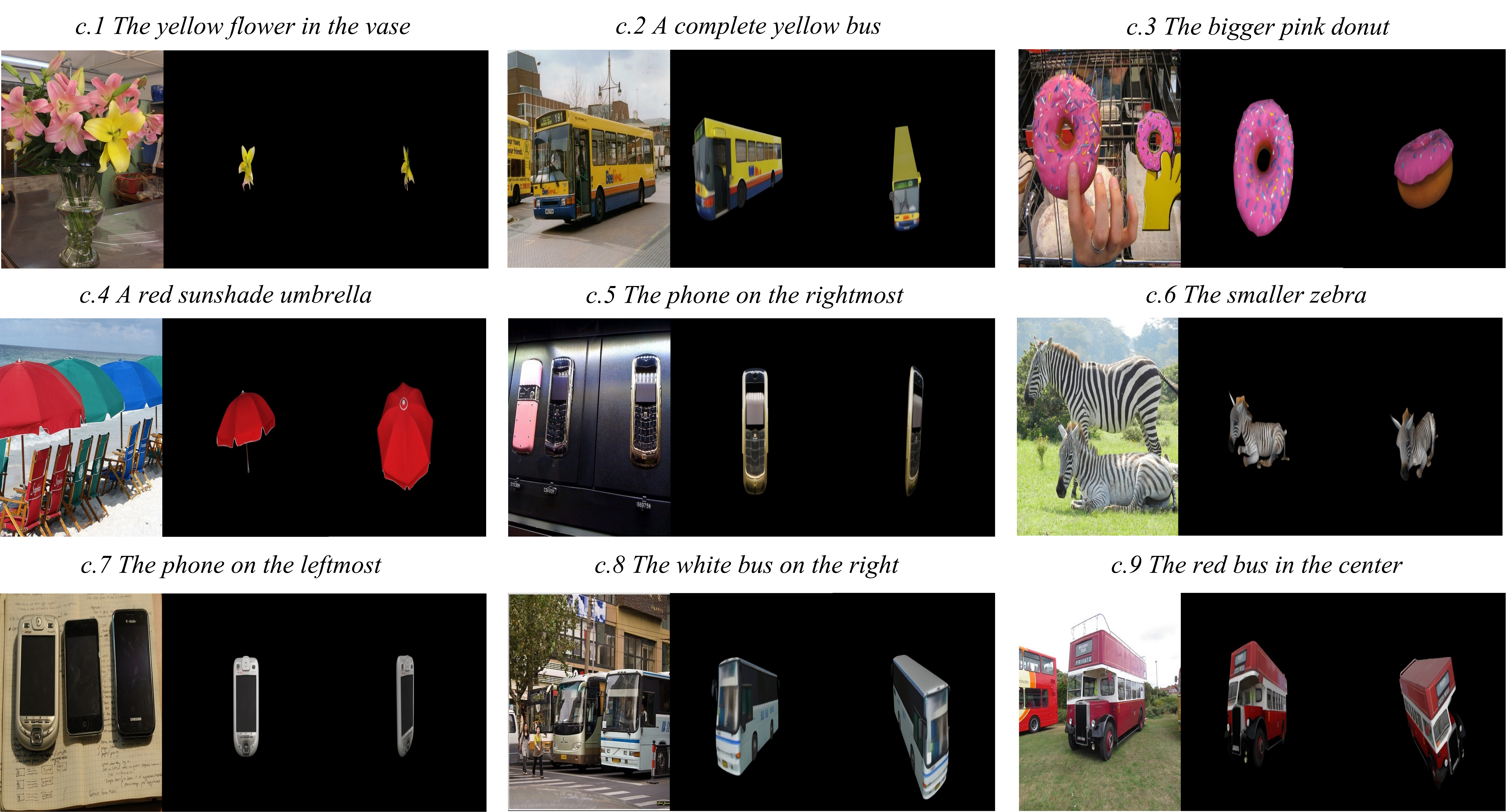}
    \caption{Case C: Referring to and reconstructing a single object across multiple instances of the same semantic class.}
    \label{fig:case_c}
  \end{center}
\end{figure*}

The proposed approach, \textbf{Ref-SAM3D}, leverages the compositional power of existing foundation models to enable text-referenced 3D reconstruction from a single RGB image. As shown in \cref{fig:pipeline}, Ref-SAM3D follows a two-stage pipeline. The first stage employs a text-to-mask segmenter (e.g., SAM3), to generate object masks based on referring expressions. These masks are used to identify and delineate specific objects in the image, which is necessary for the object reconstruction process. In the second stage, SAM3D takes the 2D image and the generated masks to reconstruct the objects in 3D. 
SAM3D internally decodes multi-modal cues into spatial representations including voxels, meshes, and Gaussian splats, along with associated geometric attributes such as rotation, translation, and scaling. These representations form a structured and comprehensive 3D characterization of each object, enabling downstream manipulation and scene integration.
Notably, the entire pipeline operates without the need for fine-tuning or joint training, using both models in their pre-trained, off-the-shelf forms. This design choice allows for a seamless and efficient process, enabling robust text-referenced 3D reconstruction with minimal customization.

Formally, given an input RGB image of a scene $\mathbf{I} \in \mathbb{R}^{H \times W \times 3}$ and a natural language expression $\mathbf{t}$ (e.g., "the blue backpack on the floor"), the first stage utilizes a mask proposer to generate a set of binary segmentation masks $\mathbf{M}_1, \mathbf{M}_2, \dots, \mathbf{M}_N \in \{0, 1\}^{H \times W}$, where each mask $\mathbf{M}_i$ corresponds to an object referred to by the text:
\begin{equation}
    \mathbf{M}_i = \mathcal{M}(\mathbf{I}, \mathbf{t}), \quad \forall i \in \{1, 2, \dots, N\}.
\end{equation}

Each mask $\mathbf{M}_i$, which delineates a target object specified by the text, is then passed along with the original image $\mathbf{I}$ to SAM3D~\cite{sam3d} to obtain the corresponding 3D reconstruction $\mathcal{R}_i$ (e.g., a mesh or point cloud) for each object:
\begin{equation}
    \mathcal{R}_i = \text{SAM3D}(\mathbf{I}, \mathbf{M}_i), \quad \forall i \in \{1, 2, \dots, N\}.
\end{equation}
The final output $\mathcal{R}$ is the collection of 3D representations $\{\mathcal{R}_1, \mathcal{R}_2, \dots, \mathcal{R}_N\}$, where each $\mathcal{R}_i$ corresponds to the 3D reconstruction of an object referred to by the input text $\mathbf{t}$.


\section{Experiments}

To evaluate the effectiveness of the proposed method in 3D referring reconstruction, we design three progressively challenging settings, each with increasing linguistic and scene complexities:

\begin{itemize}
    \item Referring to and reconstructing a single object;
    \item Referring to and reconstructing multiple objects;
    \item Referring to and reconstructing an object from multiple instances with the same semantics.
\end{itemize}

\subsection{Referring to and Reconstructing a Single Object}
In this setting, the model is given a single-view image of a 3D scene along with a natural language expression referring to an object in the scene. The model must localize the object and reconstruct its geometry in a structured 3D representation (e.g., mesh or gaussians). This task integrates 3D visual grounding and object-level 3D reconstruction. As shown in \cref{fig:case_a}, as the simplest scenario, the language expression refers to only one object.

\subsection{Referring to and Reconstructing Multiple Objects}
Building on the previous setting, we extend it to multiple objects. Here, the input language expression refers to multiple distinct objects. The model must disentangle the semantic references for each object and simultaneously reconstruct the full 3D geometry for all of them. This requires robust multi-instance grounding and reconstruction capabilities. Our qualitative results for this setting are presented in \cref{fig:case_b}.

\subsection{Referring to and Reconstructing an Object from Multi instances}
This setting presents a further challenge, where the scene contains several visually similar instances of the same object category. The language expression must use fine-grained spatial or relational cues to uniquely identify a specific instance. Successfully grounding and reconstructing the correct object in such an ambiguous context tests the model's ability to reason contextually and discriminate between instances. Results for this setting are shown in \cref{fig:case_c}.

\section{Conclusion}

In this report, we present \textbf{Ref-SAM3D}, a simple yet effective framework that extends SAM3D to support text-referenced 3D object reconstruction using only off-the-shelf foundation models, eliminating the need for fine-tuning or joint training.
Through comprehensive qualitative evaluations, we demonstrate that Ref-SAM3D achieves competitive and robust zero-shot 3D reconstruction performance across a variety of scenarios, validating the effectiveness of using natural language as an alternative reference modality.
We hope Ref-SAM3D serves as a stepping stone toward more accessible, language-driven 3D content generation and paves the way for future research at the intersection of vision, language, and 3D understanding.



\bibliography{example_paper}
\bibliographystyle{icml2026}


\end{document}